\def\printing{myecai}
\def\myecai{myecai}
\def\blind{blind}
\def\article{article}
\def\frontpage{frontpage}
\ifx\printing\myecai\documentclass{ecai2002}
   \else\ifx\printing\blind\documentclass{ecai2002}\else
   \documentclass{article}\fi\fi
\usepackage{times}
\usepackage{epic}
\usepackage{ecltree}
\usepackage{graphicx}
\usepackage{latexsym}
   
\begin{document}

\newlength{\halftextwidth}
\setlength{\halftextwidth}{0.47\textwidth}
\def\halffigsize{2.2in}
%\newlength{\thirdtextwidth}
%\setlength{\thirdtextwidth}{0.31\textwidth}
\def\thirdfigsize{1.5in}
\def\negvspace{0in}
\def\posvspace{0em}

\input epsf

%\usepackage{amssymb}

%\addtolength{\textwidth}{1.5in}
%\addtolength{\textwidth}{0.5in}
%\addtolength{\oddsidemargin}{-0.75in}
%\addtolength{\topmargin}{-0.5in}
%\addtolength{\textheight}{0.5in}

%\newcommand{\figref}[1]{Figure \ref{#1}}
%\newcommand{\tblref}[1]{Table \ref{#1}}
%\newcommand{\secref}[1]{Section \ref{#1}}
%\newcommand{\eqref}[1]{(\ref{#1})}
%\newcommand{\mygtrsim}{\gtrsim}

\newcommand{\deriv}[2]{\Delta #1_#2}

\newcommand{\tighter}[2]{\mbox{$#1 \preceq #2$}}
\newcommand{\stighter}[2]{\mbox{$#1 \prec #2$}}
\newcommand{\ac}[1]{\mbox{$ac(#1)$}}
\newcommand{\acmin}[1]{\mbox{$ac_{\min}(#1)$}}
\newcommand{\pc}[1]{\mbox{$pc(#1)$}}
\newcommand{\spc}[1]{\mbox{$spc(#1)$}}
\newcommand{\incomparable}[2]{\mbox{$#1 \sim #2$}}
\newcommand{\twodarray}[4]{\mbox{\scriptsize $\left( \hspace{-0.5em} \begin{array}{cc} #1 & #2 \\ #3 & #4 \end{array} \hspace{-0.5em} \right)$}}

\newcommand{\threedarray}[9]{\mbox{\scriptsize $\left( \hspace{-0.5em} \begin{array}{ccc} #1 & #2 & #3 \\ #4 & #5 & #6 \\ #7 & #8 & #9  \end{array} \hspace{-0.5em} \right)$}}

\renewcommand{\theenumii}{\alph{enumii}} 
\renewcommand{\theenumiii}{\roman{enumiii}} 
\newcommand{\figref}[1]{Figure \ref{#1}} 
\newcommand{\tblref}[1]{Table \ref{#1}} 
\newcommand{\size}{\mbox{$N$}} %\newcommand{\prob}[1]{\mbox{Pr\{#1\}}} 
\newcommand{\prob}[1]{\mbox{{P}r\{\scriptsize#1\}}} 
\newcommand{\probB}[1]{\mbox{{P}r\{#1\}}} 
\newcommand{\secref}[1]{Section \ref{#1}} 
\newcommand{\myfrac}[2]{(#1)/#2} 
\newcommand{\CSP}{\mbox{\sc Csp}} 
\newcommand{\Csp}{\mbox{\sc Csp}} 
\newcommand{\SAT}{\mbox{\sc Sat}}  

\def\op{\mbox{$\kappa$}} 
\def\opcrit{\mbox{$\kappa_{c}$}} 
\def\opn{\mbox{$\gamma$}} 
\newcommand{\nmpp}{\mbox{$\langle n,m,p_{1},p_{2}\rangle$}} 
\newtheorem{mytheorem}{Theorem} 
\newtheorem{mytheorem1}{Theorem} 
\newcommand{\myproof}{\noindent {\bf Proof:\ \ }} 
\newcommand{\myqed}{\mbox{QED.}} 

\title{Stochastic Constraint Programming}
\ifx\printing\blind
\author{%Toby Walsh\\ 
%Department of Computer Science \\ 
%University of York \\
%York YO10 5DD  \\
%England \\ 
%{\tt tw@cs.york.ac.uk}}
Content areas: constraint programming, constraint satisfaction\\
Tracking number: E0036\institute{Address deleted to enable blind reviewing.}}
\date{}
\else \fi

\ifx\printing\myecai
\author{
Toby Walsh\institute{Cork Constraint Computation Centre,
University College Cork, Ireland. 
email: tw@4c.ucc.ie. 
The author thanks the other
members of the APES research group, 
especially Ian Gent for his helpful discussions.
This research was supported by an EPSRC advanced research fellowship.}}
%\date{}
%\setlength\titlebox{2in}
\else \fi

\ifx\printing\article
\author{Toby Walsh\\ 
Department of Computer Science \\ 
University of York \\
York YO10 5DD  \\
England \\ 
{\tt tw@cs.york.ac.uk}}
\date{1st June 2001}
\else \fi

\ifx\printing\frontpage
\author{Toby Walsh\\ 
Department of Computer Science \\ 
University of York \\
York YO10 5DD  \\
England \\ 
{\tt tw@cs.york.ac.uk}}

\else \fi

\maketitle
\begin{abstract}
To model combinatorial decision problems involving uncertainty
and probability, we introduce stochastic
constraint programming. Stochastic constraint programs
contain both decision variables (which we can set) and
stochastic variables (which follow a probability 
distribution). They combine
together the best features of traditional constraint
satisfaction, stochastic integer programming, and
stochastic satisfiability. 
We give a semantics for stochastic constraint
programs, and propose a number of complete algorithms and
approximation procedures. % for stochastic constraint programming. 
%Using these algorithms,
%we observe phase transition behavior. % in stochastic constraint programs. 
%Interestingly, the cost of both optimization and satisfaction
%peaks in the satisfaction phase boundary. 
Finally,
we discuss a number of extensions
of stochastic constraint programming 
to relax various assumptions like
the independence between stochastic variables, 
and compare %stochastic constraint programming 
with other approaches
for decision making under uncertainty.
% like Markov
%decision problems and influence diagrams. 
\end{abstract}

\ifx\printing\frontpage
\begin{center}
\begin{tabular}{ll}
Content areas: & constraint programming, constraint satisfaction, reasoning under uncertainty\\
%Word count: & 4668 words \\
Tracking number: & 142 \\
\end{tabular}
\end{center}

\begin{quote}
This paper has not already been accepted by and is not currently under review for
a journal or another conference, nor will it be submitted for such during ECAI's
review period.
\end{quote}

\else \fi

\ifx\printing\frontpage
\eject \end{document} \else \fi

%\vspace{-2mm}
\section{Introduction}

Many %real world 
decision problems contain uncertainty. Data about events in
the past may not be known exactly due to errors in measuring or 
difficulties in sampling, whilst data about events in the future may
simply not be known with certainty. For example,
when scheduling power stations, we need to cope with uncertainty
in future energy demands. As a second example, 
nurse rostering in an accident and emergency department
requires us to anticipate variability in workload. As a final example,
when constructing a balanced bond portfolio, 
we must deal with uncertainty in the future price 
of bonds. To deal with such situations, we propose an
extension of constraint programming called {\em stochastic
constraint programming} in which we distinguish
between decision variables, which we are
free to set, and stochastic (or observed) variables, which follow some
probability distribution. 

\section{Stochastic constraint programs}

We define a number of models of stochastic constraint
programming of increasing complexity. 
In an one stage
stochastic constraint satisfaction problem (stochastic CSP), 
the decision variables
are set before the stochastic variables. 
This models situations
where we act now and observe later. 
For example, we have to decide now which nurses to
have on duty and will only later discover the actual workload. 
We can easily invert the instantiation order if
the application demands, with the stochastic variables
set before the decision variables. % are set.  
Constraints are defined (as in traditional constraint
satisfaction) by relations of allowed tuples of values.
Constraints can, however, be implemented with specialized
and efficient algorithms for consistency checking. 
The stochastic variables independently
take values with probabilities given by a %fixed 
probability distribution. We discuss later how to relax these assumptions,
and how this %model 
compares with other frameworks. %like mixed constraint satisfaction.
A one stage stochastic CSP is satisfiable iff 
there exists values for the decision 
variables so that, given random values for
the stochastic variables, 
the probability that all the constraints
are satisfied equals or exceeds a threshold $\theta$.
The probabilistic satisfaction of constraints
allows us to ignore worlds (values for the 
stochastic variables) which are rare. % to require consideration. 
Note that 
the definition reduces to that of a traditional
constraint satisfaction problem
if we have no stochastic variables and $\theta=1$. 

In a two stage
stochastic CSP,
there are two sets of decision variables, $V_{d1}$ and
$V_{d2}$, and two sets of stochastic variables,
$V_{s1}$ and $V_{s2}$. The aim is to find
values for the variables in 
$V_{d1}$, so that given random 
values for $V_{s1}$, we can find
values for $V_{d2}$, so that
given random  values for $V_{s2}$,
the probability that all the
constraints are satisfied equals or exceeds
$\theta$. Note that the values chosen for the second
set of decision variables $V_{d2}$ are conditioned
on both the values chosen for
the first set of decision variables $V_{d1}$
and on the random values given to the first
set of stochastic variables $V_{s1}$. 
This can model situations in which
items are produced and can be consumed 
or put in stock for later consumption. 
Future production then depends both
on previous production (earlier decision
variables) and on previous demand (earlier 
stochastic variables). A $m$ stage 
stochastic CSP
is defined in an analogous way to one and two
stage stochastic CSPs. 

A stochastic constraint optimization problem (stochastic COP)
is a stochastic CSP plus
a cost function defined over the decision and stochastic variables.
The aim is to find a solution that satisfies
the stochastic CSP which minimizes (or, if desired, maximizes)
the expected value of the cost function. 

\section{Production planning example}

The following %$m$ stage 
stochastic constraint program models a simple
$m$ quarter 
production planning problem. In each 
quarter, we will sell between 100 and 105 copies of a book. 
To keep customers happy, we want to satisfy 
demand in all $m$ quarters with 80\% probability. 
At the start of each quarter, we decide 
how many books to print for that quarter. 
This problem is modelled by a $m$ stage stochastic CSP.
There are $m$ decision variables, $x_i$ representing
production in each quarter. 
There are also $m$ stochastic variables,
$y_i$ representing demand in each quarter. 
These take values between $100$ and $105$ with
equal probability.
There is a constraint to ensure 1st
quarter production
meets 1st quarter demand:
$$ x_1 \geq y_1$$
There is also a constraint to ensure 2nd
quarter production
meets 2nd quarter demand plus any unsatisfied demand %from the 1st quarter 
or less any stock: % carried forward:
$$ x_2 \geq y_2 + (y_1 - x_1)$$
And there is a constraint 
to ensure $j$th quarter production ($j \geq 2$) 
meets $j$th quarter demand plus any unsatisfied demand
%from earlier quarters 
or less any stock: % carried forward:
$$ x_j \geq  y_j + \sum_{i=1}^{j-1} (y_i - x_i)$$
We must satisfy these $m$ constraints with 
a threshold probability $\theta = 0.8$. 
This stochastic CSP has a number of solutions
including $x_i=105$ for each $i$ (i.e. always produce
as many books as the maximum demand).
However, this solution will tend to produce books surplus
to demand which is undesirable. 

Suppose storing surplus book costs \$1 per quarter.
We can define a $m$ stage stochastic COP based
on this stochastic CSP in which we additionally miminize the 
expected cost of storing surplus books. 
As the number of surplus books
in the $j$th quarter is $\min( \sum_{i=1}^j  x_i - y_i, 0)$,
we have a cost function 
over all quarters of:
$$\sum_{j=1}^m \min(\sum_{i=1}^j x_i - y_i, 0)$$
Note that a solution to a stochastic CSP or COP
defines how to set later
decision variables given the values for earlier stochastic
and decision variables. 

\section{Semantics}

A stochastic constraint satisfaction problem is a 6-tuple 
$\langle V, S, D, P, C, \theta \rangle$ 
where $V$ is a list of variables,
$S$ is the subset of $V$ which are stochastic varibles,
$D$ is a mapping from $V$ to domains, 
$P$ is a mapping from $S$ to probability distributions for the domains,
$C$ is a set of constraints over $V$,
and $\theta$ is a threshold probability in the interval $[0,1]$.
Constraints are defined %(as in traditional constraint satisfaction) 
by a set of variables and 
a relation giving the allowed tuples of values. % for these variables.
Variables are set in the order
in which they appear in $V$. Thus,
in an one stage stochastic CSP, $V$ contains 
the decision variables
and then the stochastic variables. 
In a two stage stochastic CSP, $V$ contains the
first set of decision variables, the
first set of stochastic variables, then
the second set of decision variables, and finally the
second set of stochastic variables. 

A {\bf policy} is a tree with nodes labelled with
variables, starting with the first variable in $V$ labelling the
root, and ending with the last variable in $V$ labelling the nodes
directly above the leaves. Nodes labelled with 
decision variables have a single child,
whilst nodes labelled with 
stochastic variables have one child for every possible value. 
Edges in the tree are labelled with values assigned to
the variable labelling the node above. 
Leaf nodes are labelled with 1
if the assignment of values to
variables along the path to the root satisfies all the constraints,
and 0 otherwise. 
Each leaf
node corresponds to a possible world and has an associated
probability; if $s_i$ is the $i$th stochastic
variable on a path to the root, $d_i$ is the
value given to $s_i$ on this path,  (i.e. the label of the following edge), 
and $prob(s_i=d_i)$
is the probability that $s_i=d_i$, then the probability
of this world is simply $\prod_i prob(s_i=d_i)$. 
We define the {\bf satisfaction} of a policy 
as the sum of the leaf values weighted
by their probabilities. A policy {\bf satisfies} the constraints
iff its satisfaction is at least $\theta$. 
A stochastic %constraint satisfaction problem 
CSP 
is  {\bf satisfiable} iff there is a policy
which {satisfies} the constraints. 
%When $S=\{\}$ and $\theta=1$, this reduces
%to the traditional definition of constraint satisfaction.
The {\bf optimal satisfaction} of a stochastic CSP is
the maximum satisfaction of all policies. 
For a stochastic COP, the {\bf expected value}
of a policy is the sum of the objective valuations
of each leaf node weighted by their probabilities.
A policy is {\bf optimal} if it satisfies
the constraints and maximizes (or, if desired, minimizes)
the expected value.

Consider again the production planning problem
and a two-quarter policy that 
sets $x_1=104$ and if 
$y_1 > 100$ then $x_2=y_1+1$
else $y_1=100$ and $x_2=100$. 
We can represent this policy by the 
following (partial) tree:

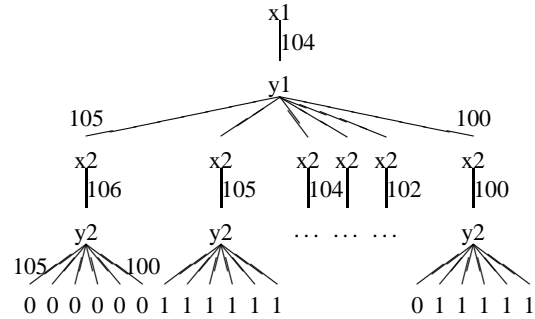
\begin{center}
\begin{bundle}{x1}
        \chunk[\ \ \ \ \ \  104]{\begin{bundle}{y1}
                      \chunk[105]{
\begin{bundle}{x2}
        \chunk[\ \ \ \ \ \  106]{\begin{bundle}{y2}
                      \chunk[105]{0}
                      \chunk[]{0}
                      \chunk[]{0}
                      \chunk[]{0}
                      \chunk[]{0}
                      \chunk[100]{0}
                    \end{bundle}}
\end{bundle}

}
                      \chunk[]{\begin{bundle}{x2}
        \chunk[\ \ \ \ \ \  105]{\begin{bundle}{y2}
                      \chunk[]{1}
                      \chunk[]{1}
                      \chunk[]{1}
                      \chunk[]{1}
                      \chunk[]{1}
                      \chunk[]{1}
                    \end{bundle}}
\end{bundle}

}
                      \chunk[]{

\begin{bundle}{x2}
        \chunk[\ \ \ \ \ \  104]{\ldots}
\end{bundle}
}
                      \chunk[]{
\begin{bundle}{x2}
        \chunk[]{\ldots}
\end{bundle}
}
                      \chunk[]{
\begin{bundle}{x2}
        \chunk[\ \ \ \ \ \ 102]{\ldots}
\end{bundle}

}

                      \chunk[100]{
\begin{bundle}{x2}
        \chunk[\ \ \ \ \ \  100]{\begin{bundle}{y2}
                      \chunk[]{0}
                      \chunk[]{1}
                      \chunk[]{1}
                      \chunk[]{1}
                      \chunk[]{1}
                      \chunk[]{1}
                    \end{bundle}}
\end{bundle}
}
                    \end{bundle}}
\end{bundle}
\end{center}

By definition, each of the leaf nodes in this
tree is equally probable. 
There are $6^2$ leaf nodes,
of which only $7$ are labelled 0.
Hence, this policy's satisfaction is
$(36-7)/36$, 
and the policy satisfies the constraints
as this just exceeds $\theta=0.8$.

\section{Complexity}

Constraint satisfaction is NP-complete in general.
Not surprisingly, stochastic constraint satisfaction moves
us up the complexity hierarchy. It may therefore be useful
for modelling problems like reasoning under uncertainty
which lie in these higher complexity classes. 
We show how a number of satisfiability 
problems in these higher complexity
classes reduce to stochastic constraint satisfaction. 
In each case, the reduction is very immediate. 
Note that each reduction can be restricted to stochastic
CSPs on binary constraints using a hidden variable
encoding to map non-binary constraints
to binary constraints. The hidden variables are added to the last
stage of the stochastic CSP. 

PP, or probabilistic polynomial time is characterized
by the PP-complete problem, {\sc Majsat} which
decides if at least half the assignments to a set of Boolean
variables satisfy a clausal formula. This can be 
reduced to a one stage stochastic CSP in
which there are no decision variables, the
stochastic variables are Boolean, the constraints
are the clauses, the two truth values for the stochastic variables 
are equally likely
and the threshold probability $\theta = 0.5$. 
A number of other reasoning problems like 
plan evaluation in probabilistic domains
are PP-complete.

NP$\mbox{}^{\rm PP}$ is the class of problems that
can be solved by non-deterministic guessing a solution
in polynomial time (NP) and then verifying this
in probabilistic polynomial time (PP). Given a
clausal formula, {\sc E-Majsat} 
is the problem of deciding if there exists
an assignment for a set of Boolean variables
so that, given randomized choices of values for
the other variables, the formula is satisfiable
with probability at least equal to some threshold $\theta$
\cite{littman97e}.
This can be reduced very immediately to an one stage stochastic CSP.
A number of other reasoning problems like 
finding optimal size-bounded plans in uncertain domains
are NP$\mbox{}^{\rm PP}$-complete.

{\sc Pspace} is the class of problems that
can be solved in polynomial space. 
Note that NP $\subseteq$
PP $\subseteq$
NP$\mbox{}^{\rm PP}$ $\subseteq$
{\sc Pspace}. 
{\sc Ssat}, or stochastic
satisfiability is an example of a {\sc Pspace}-complete problem.
In {\sc Ssat}, we have a clausal formula with
$m$ alternating decision and stochastic variables, and
must decide if the formula is satisfiable 
with probability at least equal to some threshold $\theta$.
This can be immediately reduced to a $m$ stage stochastic CSP. 
A number of other reasoning problems like 
propositional STRIPS planning 
are {\sc Pspace}-complete.

\section{Complete algorithms}

We present a backtracking algorithm for solving stochastic
CSPs, which
is then extended to a forward checking procedure.
 
\subsection{Backtracking}

We assume that variables are instantiated
in order. However, if decision variables
occur together, they can be instantiated in any order.
A branching heuristic like fail
first may therefore be used to order decision
variables which occur together.
On meeting a decision variable, the backtracking (BT) algorithm
tries each value in its domain in turn.
The maximum value is returned to the previous
recursive call. 
On meeting a stochastic variable, we
try each value in turn, and
returns the sum of the all answers to the subproblems
weighted by the probabilities of their occurrence. 
At any time, if instantiating a decision
or stochastic variable breaks a constraint,
we return 0. If we manage to instantiate all the variables
without breaking any constraint, we return 1.
%The algorithm can be trivially adapted to
%record the optimal policy. 
\begin{figure}[htb]
{\scriptsize
\begin{tabbing}
{\bf procedure} BT($i$,$\theta_l$,$\theta_h$) \\
~ ~ \= {\bf if} $i > n$ {\bf then} return 1 \\
      \> $\theta := 0$ \\
      \> $q := 1$ \\  
      \> {\bf for each} $d_j \in D(x_i)$ \\
      \>  ~ ~ \= {\bf if} $x_i \in S$ {\bf then} \\
      \>      \> ~ ~ \= $p := $ prob($x_i \rightarrow d_j$) \\
      \>      \>       \> $q := q - p$ \\
      \>      \>       \> {\bf if} $consistent(x_i \rightarrow d_j$) {\bf then} \\
      \>      \>       \> ~ ~ \= $\theta := \theta$ + $p \times$ BT($i+1$,$\frac{\theta_l-\theta-q}{p}$,$\frac{\theta_h-\theta}{p}$) \\
      \>      \>       \>       \> {\bf if} $\theta > \theta_h$ {\bf then} return $\theta$ \\

      \>      \>       \> {\bf if} $\theta+q < \theta_l$ {\bf then} return $\theta$ \\
      \>      \> {\bf else} \\
      \>       \>  \> {\bf if} $consistent(x_i \rightarrow d_j$) {\bf then} \\
      \>       \>  \> \> $\theta := \max$($\theta$,BT($i+1$,$\max(\theta,\theta_l)$,$\theta_h$)) \\
      \>       \>  \> \> {\bf if} $\theta > \theta_h$ {\bf then} return $\theta$ \\
      \> return $\theta$ 
\end{tabbing}}
\vspace{-1em}
\caption{The backtracking (BT) algorithm 
%{\scriptsize {\bf Figure 1.} The backtracking (BT) algorithm 
%for stochastic 
%constraint satisfaction problems.
%CSPs.
%The algorithm 
is called with the search
depth, $i$ and with %upper and lower 
bounds, $\theta_h$ and $\theta_l$.
If the optimal satisfaction lies between these
bounds, BT returns the exact satisfaction. 
If the optimal satisfaction is $\theta_h$ or
more, BT returns a value greater than or equal to $\theta_h$.
If the optimal satisfaction is $\theta_l$ or
less, BT returns a value less than or equal to $\theta_l$.
$S$ is the set of stochastic variables.
}
\end{figure}

%As in the Davis-Putnam like algorithm for stochastic
%satisfiability \cite{littman1},
Upper and lower bounds, $\theta_h$ and $\theta_l$ are used to prune search.
By setting $\theta_l=\theta_h=\theta$, we can
determine if the optimal satisfaction is
at least $\theta$. 
Alternatively, by setting $\theta_l=0$ and $\theta_h=1$, we can
determine the optimal satisfaction.
The calculation of upper and lower bounds in recursive calls
requires some explanation. 
Suppose that the current assignment to
a stochastic variable returns a satisfaction of
$\theta_0$. We can safely ignore other values 
for this stochastic variable if
$\theta + p \times \theta_0 \geq \theta_h$.
That is, if $\theta_0 \geq \frac{\theta_h-\theta}{p}$. This 
gives the upper bound in the recursive call to BT
on a stochastic variable.
Alternatively, 
we cannot hope to satisfy the
constraints adequately if
$\theta + p \times \theta_0 + q \leq \theta_l$
as $q$ is the maximum that the remaining values can
contribute to the satisfaction. 
That is, if $\theta_0 \leq \frac{\theta_l-\theta-q}{p}$. This 
gives the lower bound in the recursive call to BT
on a stochastic variable.
Finally, suppose that the 
current assignment to
a decision variable returns a satisfaction of
$\theta$. If this is more that $\theta_l$,
then any other values must exceed $\theta$
to be part of a better policy. Hence, we can
replace the lower bound in the recursive call to BT
on a decision variable by $\max(\theta,\theta_l)$. 
Because of these bounds, value ordering heuristics can reduce
search. For decision variables, we should
choose values that are likely to return the optimal satisfaction.
For stochastic variables, we should
choose values that are more likely. 

\subsection{Forward checking}

The Forward Checking (FC) procedure is based
on the BT algorithm. 
On instantiating a decision or stochastic variable, the FC algorithm
checks forward and prunes values
from the domains of future decision and stochastic variables
which break constraints. Checking forwards fails if a
stochastic or decision variable has a domain wipeout (dwo),
or if a stochastic variable has so many values removed
that we cannot hope to satisfy the constraints. 
As in the regular forward checking algorithm, we can use
an 2-dimensional array, $prune(i,j)$ to record the depth at which 
the value $d_j$ for the variable $x_i$ is removed by forward checking.
This is used to restore values on backtracking. In addition, each 
stochastic variable, $x_i$ has an upper bound, $q_i$
on the probability that the values left in its
domain can contribute to a solution. When forward checking removes
some value, $d_j$ from $x_i$, we reduce $q_i$ by $prob(x_i \rightarrow d_j)$,
the probability that $x_i$ takes the value $d_j$. 
This reduction on $q_j$ 
is undone on backtracking. 
If forward checking ever 
reduces $q_i$ to less than $\theta_l$, we %  immediately 
backtrack as it is %now 
impossible to set $x_i$ and satisfy the constraints
adequately.
\begin{figure}[htbp]
{\scriptsize
\begin{tabbing}
{\bf procedure} FC($i$,$\theta_l$,$\theta_h$) \\
~ ~ \= {\bf if} $i > n$ {\bf then} return 1 \\
      \> $\theta := 0$ \\
      \> {\bf for each} $d_j \in D(x_i)$ \\
      \> ~ ~ \= {\bf if} $prune(i,j)=0$ {\bf then} \\
      \>       \>  ~ ~ \= {\bf if} check($x_i \rightarrow d_j$,$\theta_l$) {\bf then} \\
      \>      \>       \> ~ ~ \= {\bf if} $x_i \in S$ {\bf then} \\
      \>      \>       \>       \> ~ ~ \= $p := $ prob($x_i \rightarrow d_j$) \\
      \>      \>       \>       \>       \> $q_i := q_i - p$ \\
      \>      \>       \>       \>       \> $\theta := \theta + p \times $ FC($i+1$,$\frac{\theta_l-\theta-q_i}{p}$,$\frac{\theta_h-\theta}{p}$) \\
      \>      \>       \>       \>       \> restore($i$) \\
      \>      \>       \>       \>       \> {\bf if} $\theta+q_i < \theta_l$ {\bf then} return $\theta$ \\
      \>      \>       \>       \>       \> {\bf if} $\theta > \theta_h$ {\bf then} return $\theta$ \\
      \>      \>      \>       \> {\bf else} \\
      \>      \>      \>   \> \> $\theta := \max$($\theta$,FC($i+1$,$\max(\theta,\theta_l)$,$\theta_h$)) \\      \>      \>      \>   \> \> restore($i$) \\
      \>      \>      \>   \> \> {\bf if} $\theta > \theta_h$ {\bf then} return $\theta$ \\
      \>      \>   \> {\bf else} restore($i$) \\
      \>      return $\theta$ \\      
~ \\
{\bf procedure} check($x_i \rightarrow d_j$,$\theta_l$) \\
      \> {\bf for} $k=i+1$ {\bf to} $n$ \\
      \>     \> dwo := true \\
      \>     \> {\bf for} $d_l \in D(x_k)$ \\
      \>     \>        \> {\bf if} $prune(k,l)=0$ {\bf then} \\
      \>     \>        \>   \> {\bf if} inconsistent($x_i \rightarrow d_j$,$x_k \rightarrow d_l$) {\bf then} \\
      \>     \>        \>   \>       \> $prune(k,l) := i$ \\
      \>     \>        \>   \>       \> {\bf if} $x_k \in S$ {\bf then} \\
      \>     \>        \>   \>       \> ~ ~ \= $q_k := q_k - $ prob($x_k \rightarrow d_l$) \\
      \>     \>        \>   \>       \>       \> {\bf if} $q_k < \theta_l$ {\bf then} return false \\
      \>     \>        \>   \> {\bf else} dwo := false \\
      \>     \> {\bf if} dwo {\bf then} return false \\
      \> return true \\
~ \\
{\bf procedure} restore($i$) \\
     \> {\bf for} $j=i+1$ {\bf to} $n$ \\
     \>      \> {\bf for} $d_k \in D(x_j)$ \\
     \>      \>       \> {\bf if} $prune(j,k)=i$ {\bf then} \\
     \>      \>       \>   \> $prune(j,k)=0$ \\
     \>     \>        \>   \> {\bf if} $x_j \in S$ {\bf then} $q_j := q_j + $ prob($x_j \rightarrow d_k$) 
\end{tabbing}
}
\vspace{-1em}
\caption{The forward checking
%{\scriptsize {\bf Figure 2.} The forward checking
(FC) algorithm
% for stochastic %constraint satisfaction problems. CSPs.
%The algorithm 
is called with the search
depth, $i$ and %upper and lower 
bounds, $\theta_h$ and $\theta_l$.
If the maximum satisfaction of all policies lies between these
bounds, FC returns the exact maximum satisfaction. 
If the maximum satisfaction of all policies is $\theta_h$ or
more, FC returns a value greater than or equal to $\theta_h$.
If the minimum satisfaction of all policies is $\theta_l$ or
less, FC returns a value less than or equal to $\theta_l$.
$S$ is the set of stochastic variables. 
The array $q_i$ is an upper bound on the probability that
the stochastic variable $x_i$ satisfies the constraints
and is initially set to 1, whilst $prune(i,d)$ is the
depth at which the value $d$ is pruned from $x_i$ and is 
initially set to $0$ which indicates that the value is not
yet pruned. 
}
\end{figure}

\section{Experimental evaluation}

We implemented the BT and FC algorithms in Common Lisp
and ran them on the production planning problem
given in Section 3, as well as on a range of 
randomly generated problems. For the production plannng
problem, we use a simple heuristic which 
orders values for the decision variables by their size. This
will tend to keep stock levels low. It will also 
ensure that the worlds in which we fail to satisfy
the demand constraints are those where demand is much
higher than average.
Results are given in Table 1
with the threshold for satisfiability
$\theta$ set to 0.8. 
Similar
results are obtained for other non-zero $\theta$. 
Surprisingly, performance was relatively
insensitive to the precise value of $\theta$
used.

\begin{table}
\begin{center}
\begin{tabular}{|r||rr|rr|} \hline
Number of &  \multicolumn{2}{c|}{BT} & \multicolumn{2}{c|}{FC}  \\
quarters & nodes & CPU/sec & nodes & CPU/sec \\ \hline \hline
1 & 28 & 0.01 & 10 & 0.01 \\
2 & 650 & 0.09 & 148 & 0.03 \\
3 & 17,190 & 2.72 & 3,604 & 0.76 \\
4 & 510,346 & 83.81 & 95,570 & 19.07 \\
5 & 15,994,856 & 3,245.99 & 2,616,858 & 509.95 \\ \hline
\end{tabular}
\caption{Backtracking (BT) and forward checking
(FC) algorithms on the production planning problem
from Section 3.
CPU times are for a Common Lisp implementation
running under Linux on an ancient 133MHz Pentium,
whilst ``nodes'' are the number of nodes visited
in the and/or search tree. 
%As in the text, the threshold for satisfiability
%$\theta$ is set to 0.8. 
}
\end{center}
\end{table}

The performance advantage of the FC algorithm
over the BT algorithm increases as the 
stochastic CSP increases in size.
On the largest problem in Table 1, the FC algorithm visits approximately
1/6th the search nodes in roughly 1/6th the CPU time. 
This is in line with
our results on random problems, where
the FC algorithm
is often an order of magnitude faster
than the BT algorithm. 
Even larger gains can be expected on
problems in which constraints apply
to variables which are set far apart in the
search tree.
On such problems, forward checking will prune domains
far down the search tree, thereby avoiding deep
backtracks. 

Our results show that the FC algorithm clearly
dominates the BT algorithm. 
Consistency testing and domain
pruning ensures that it
only visits a small fraction of the possible worlds.
Further performance gains could be obtained
by more powerful constraint propagation,
more intelligent backtracking, and
more sophisticated branching heuristics.
These are all areas for future work.

\section{Approximation procedures}

There are a number of methods for
approximating the answer
to a stochastic constraint program.
For example, we can replace the stochastic variables
in a stochastic % constraint satisfaction problem
CSP 
by their most probable values (or in ordered domains like
integers by their median or integer mean values), 
and then solve (or approximate the answer to) the 
resulting traditional constraint satisfaction 
problem. Similarly, we can estimate the optimal solution 
for a stochastic %constraint optimization problem 
COP by 
replacing the stochastic variables by
their most probable values and then finding (or approximating 
the answer to) the 
resulting traditional constraint optimization problem. 
We can also use Monte Carlo sampling to test a subset
of the possible worlds. For example, we can
randomly generate values for 
the stochastic variables according to their probability
distribution. 
If the fraction 
of the resulting constraint
satisfaction problems that are satisfiable
is at least equal to the threshold $\theta$, then
the original stochastic constraint satisfaction
problem is likely to be satisfiable. 
It would also be interesting
to develop local search procedures like GSAT and WalkSAT \cite{selman1,selman-aaai94}
which explore the ``policy space'' of stochastic
constraint programs.

\section{Extensions}

We have assumed that stochastic variables are 
independent. There are problems which may
require us to relax this restriction. 
For example,
a stochastic variable representing electricity demand
may depend on a stochastic variable representing
temperature. It may therefore be useful
to combine stochastic programming with techniques
like Bayes networks which allow for
conditional dependencies to be efficiently
and effectively represented. 
An alternative solution is to replace the 
dependent stochastic variables
by a single stochastic variable whose 
domain is the product space of the dependent
variables. This is only feasible
when there are a small number of dependent 
variables with small domains. 

We have also assumed that the
probability distribution of stochastic variables is
fixed, and does not depend
on earlier decision variables.
Again, there are problems which may require
us to relax this restriction. For example,
the decision variable representing price 
may influence a stochastic variable representing demand. 
A solution may again be to combine stochastic programming with techniques
like Bayes networks. 
We have also assumed that the probability distribution
is known in advance. It would be interesting
to explore methods for estimating it based on
observation. 

Finally, we have assumed that all variable domains are
finite. There are problems which may require us to
relax this restriction. For example, in scheduling
power stations, we may use 0/1 decision variables 
to model whether a power station runs or not,
but have continuous (observed) variables to
model future electricity demands. A continuous
probability density function could be associated with
these variables. Similarly, 
a continuous decision variable could be useful
to model the power output. % of the power stations. 
Interval reasoning
techniques could be extended to deal with 
such variables.

\section{Related work in decision making under uncertainty}

Stochastic constraint programs are closely
related to Markov decision problems (MDPs).
An MDP model consists of a set of states, a set
of actions, a state transition function which
gives the probability of moving between two states
as a result of a given action, and a reward function.
A solution to an MDP is a policy, which specifies 
the best action to take in each possible state.
MDPs 
These have been very influential in AI of late for
dealing with situations involving reasoning under 
uncertainty \cite{puterman1}. 
Stochastic constraint programs can model
problems which lack the Markov property that
the next state and reward depend only on the previous
state and action taken.
To represent a stochastic constraint program
in which the current decision depends on all 
earlier decisions would require an MDP
with an exponential number of states. 
Stochastic constraint optimization can also be used to
model more complex reward functions than the
(discounted) sum of individual rewards.

Stochastic constraint programs are also closely 
related to influence diagrams.
Influence diagrams are Bayesian networks in which
the chance nodes are augmented with
decision and utility nodes \cite{oliver1}. 
The usual aim is to maximize the sum of the expected
utilities. Chance nodes in an influence
diagram correspond to stochastic variables in a stochastic
constraint program,
whilst decision nodes correspond to decision variables.
The utility nodes correspond to the cost function
in a stochastic constraint optimization problem.
It would therefore be relatively straightforward
to map stochastic constraint programs into
influence diagrams. 
However, reasoning about
stochastic constraint programs is likely
to be easier than about influence diagrams.
First, the probabilistic aspect of a stochastic constraint
program is simple and decomposable as there
are only unary marginal probabilities. 
Second, the dependencies between decision variables and
stochastic variables are represented by declarative constraints.
We can therefore borrow 
from traditional constraint satisfaction and optimization
powerful algorithmic
techniques like branch and bound, constraint propagation
and nogood recording.
As a result, if a problem can be modelled within the
more restricted format of a stochastic constraint program,
we hope to be able to reason about it more efficiently. 

\section{Related work in constraints}

Stochastic constraint programming is inspired
by both stochastic integer programming and
stochastic satisfiability \cite{littman1}. 
It shares the
advantages that constraint programming has
over integer programming (e.g. non-linear
constraints, and constraint propagation).
It also shares the
advantages that constraint programming has
over satisfiability (e.g. global 
and arithmetic constraints, and more compact models).

Mixed constraint satisfaction \cite{fargier2} is 
closely related to one stage stochastic constraint
satisfaction. In a mixed CSP, the decision variables
are set after the stochastic variables are given 
random values. In addition, the random values are
chosen uniformly. In the case of full observability,
the aim is to find conditional values
for the decision variables in a mixed CSP 
so that we satisfy all possible
worlds. In the case of no observability, 
the aim is to find values
for the decision variables in a mixed CSP 
so that we satisfy as many
possible worlds.
An earlier constraint satisfaction model
for decision making under uncertainty \cite{fargier3} 
also included a probability distribution over
the space of possible worlds. 

Constraint satisfaction has been extended to include
probabilistic preferences on the values assigned to
variables \cite{shazeer1}. Associated with the values for
each variable is a probability distribution. A ``best'' solution
to the constraint satisfaction problem is then found. 
This may be the maximum probability solution (which satisfies
the constraints and is most probable), or the maximum expected
overlap solution (which is most like the true solution). 
The latter %maximum expected solution 
can be viewed as the solution 
which has the maximum expected overlap with one
generated at random using the probability distribution.
The maximum expected overlap solution could be 
found by solving a suitable one stage stochastic
constraint optimization problem. 

Branching constraint satisfaction \cite{fowler1} models
problems in which there is
uncertainty in the number of variables. For example,
we can model a nurse rostering problem
by assigning shifts to nurses. Branching constraint
satisfaction then allows us to deal with the uncertainty
in which nurses are available for duty. 
We can represent such problems with a stochastic
%constraint program 
CSP with a stochastic 0/1 variable
for each nurse representing their availability. % for work. 

A number of extensions of the traditional constraint
satisfaction problem model constraints that
are uncertain, probabilistic or not necessarily satisfied. 
For example, in partial constraint satisfaction
we maximize the number of constraints satisfied \cite{freuder5}. 
As a second example, in probabilistic constraint satisfaction
each constraint has a certain probability independent of
all other probabilities of being part of the problem \cite{fargier1}.
As a third example, both valued and semi-ring based constraint satisfaction
\cite{schiex5} generalizes probabilistic constraint satisfaction
as well as a number of other frameworks. In semi-ring
based constraint satisfaction, a value is 
associated with each tuple in a constraint, 
whilst in valued constraint satisfaction, a 
value is associated with each constraint.
However, none of these extensions deal with variables
that may have uncertain or probabilistic values. 
Indeed, stochastic constraint
programming can easily be combined with most
of these techniques. 
For example, we can define stochastic partial 
constraint satisfaction in which we maximize
the number of satisfied constraints,
or stochastic probabilistic
constraint satisfaction in which 
each constraint has an associated probability
of being in the problem.

%\vspace{-2mm}
\section{Conclusions} 

We have proposed stochastic constraint programming,
an extension of constraint programming to deal with
both decision variables (which we can set) and
stochastic variables (which follow some probability 
distribution). This framework is designed to take
advantage of the best features of traditional constraint
satisfaction, stochastic integer programming, and
stochastic satisfiability. It can be used to model 
a wide variety of decision problems involving uncertainty
and probability. 
We have given a semantics for stochastic constraint
programs based upon policies. These determine how
decision variables are set depending on
earlier decision and stochastic variables.
We have proposed a number of complete algorithms and
approximation procedures for stochastic
constraint programming. 
%Using these algorithms,
%we have observed phase transition behavior in
%stochastic constraint programs. 
%Interestingly, the cost of both optimization and satisfaction
%peaks along the satisfaction phase boundary. 
Finally,
we have discussed a number of extensions
of stochastic constraint programming to relax assumptions like
the independence between stochastic variables, and compared
it %stochastic constraint programming
with other approaches
for decision making under uncertainty like Markov
decision problems and influence diagrams.

\ifx\printing\article
\vspace{-2mm}
\section*{Acknowledgements} 
The author is an EPSRC advanced research fellow. 
He thanks the other
members of the APES research group (http://apes.cs.strath.ac.uk/),
especially Ian Gent for his helpful discussions.
\fi

\bibliographystyle{ecai2002}

%\bibliographystyle{plain}
%\bibliographystyle{alpha}
%\bibliography{/home/s5/tw/biblio/a-z,/home/s5/tw/biblio/pub}
%\bibliography{/n/endjinn/u6/tw/biblio/a-z,/n/endjinn/u6/tw/biblio/pub}
%\bibliography{/usr/tw/biblio/a-z,/usr/tw/biblio/pub}
%\bibliography{/home/arp/disk1/tw/biblio/a-z,/home/arp/disk1/tw/biblio/pub}
%\bibliography{/u6/tw/biblio/a-z,/u6/tw/biblio/pub}

\newcommand{\etalchar}[1]{$^{#1}$}
\begin{thebibliography}{FLMCS95}

\bibitem[BFM{\etalchar{+}}96]{schiex5}
S.~Bistarelli, H.~Fargier, U.~Montanari, F.~Rossi, T.~Schiex, and
  G.~Verfaillie.
\newblock Semi-ring based {CSPs} and valued {CSPs}: Basic properties and
  comparison.
\newblock %In M.~Jample, E.~Freuder, and M.~Maher, editors, 
{\em   Over-Constrained Systems}, pages 111--150. Springer-Verlag, 1996.
\newblock LNCS 1106.

\bibitem[FB00]{fowler1}
D.~Fowler and K.~Brown.
\newblock Branching constraint satisfaction problems for solutions robust under
  likely changes.
\newblock In {\em Proc. of 6th Int. Conf. on Principles and
  Practices of Constraint Programming}. Springer-Verlag, 2000.

\bibitem[FL93]{fargier1}
H.~Fargier and J.~Lang.
\newblock Uncertainty in constraint satisfaction problems: a probabilistic
  approac h.
\newblock In {\em Proc. of {ECSQARU}}. Springer-Verlag, 1993.
\newblock LNCS 747.

\bibitem[FLMCS95]{fargier3}
H.~Fargier, J.~Lang, R.~Martin-Clouaire, and T.~Schiex.
\newblock A constraint satisfaction framework for decision under uncertainty.
\newblock In {\em Proc. of the 11th Int. Conf. on
  Uncertainty in AI}, 1995.

\bibitem[FLS96]{fargier2}
H.~Fargier, J.~Lang, and T.~Schiex.
\newblock Mixed constraint satisfaction: a framework for decision problems
  under in complete information.
\newblock In {\em Proc. of the 13th Nat. Conf. on AI}. 1996.

\bibitem[FW92]{freuder5}
E.~Freuder and R.~Wallace.
\newblock Partial constraint satisfaction.
\newblock {\em Artificial Intelligence}, 58:21--70, 1992.

\bibitem[LGM98]{littman97e}
Michael~L. Littman, Judy Goldsmith, and Martin Mundhenk.
\newblock The computational complexity of probabilistic plan existence and
  evaluation.
\newblock {\em J. of AI Research}, 9:1--36, 1998.

\bibitem[LMP00]{littman1}
M.L. Littman, S.M. Majercik, and T.~Pitassi.
\newblock Stochastic {Boolean} satisfiability.
\newblock {\em J. of Automated Reasoning}, 2000.

\bibitem[OS90]{oliver1}
R.M. Oliver and J.Q. Smith.
\newblock {\em Influence Diagrams, Belief Nets and Decision Analysis}.
\newblock John Wiley and Sons, 1990.

\bibitem[Put94]{puterman1}
M.L. Puterman.
\newblock {\em Markov decision processes: discrete stochastic dynamic
  programming}.
\newblock John Wiley and Sons, 1994.

\bibitem[SKC94]{selman-aaai94}
B.~Selman, H.~Kautz, and B.~Cohen.
\newblock {Noise} {Strategies} for {Improving} {Local} {Search}.
\newblock In {\em Proc. of the 12th Nat. Conf. on AI}, pages
  337--343. 1994.

\bibitem[SLK99]{shazeer1}
N.~Shazeer, M.~Littman, and G.~Keim.
\newblock Constraint satisfaction with probabilistic preferences on variable
  values.
\newblock In {\em Proc. of the 16th Nat. Conf. on AI}. 1999.

\bibitem[SLM92]{selman1}
B.~Selman, H.~Levesque, and D.~Mitchell.
\newblock A {New} {Method} for {Solving} {Hard} {Satisfiability} {Problems}.
\newblock In {\em Proc. of the 10th Nat. Conf. on AI}, pages
  440--446. 1992.

\end{thebibliography}
\newcommand{\etalchar}[1]{$^{#1}$}

\end{document}